\documentclass{article}
\usepackage[T1]{fontenc}
\usepackage{iclr2027_conference,times}
\usepackage{multirow}
\usepackage{wrapfig}
\usepackage{float}
\usepackage{placeins}
\usepackage{dblfloatfix}
\usepackage{algorithm}
\usepackage{algpseudocode}
\usepackage{amssymb}

\usepackage{amsmath,amsfonts,bm}

\def\eqref#1{equation~\ref{#1}}

\def\1{\bm{1}}

\def\vs{{\bm{s}}}

\DeclareMathAlphabet{\mathsfit}{\encodingdefault}{\sfdefault}{m}{sl}
\SetMathAlphabet{\mathsfit}{bold}{\encodingdefault}{\sfdefault}{bx}{n}

\usepackage{amsmath}
\usepackage{booktabs}
\usepackage[most]{tcolorbox} 
\usepackage{graphicx}
\usepackage{hyperref}
\usepackage{url}

\title{Learning Teacher Output Preferences from Final Responses for Reasoning Distillation}
\author{
Hongli Shen$^{1,\dagger}$ \quad
Shaopeng Fu$^{1}$ \quad
Qinbo Zhang$^{2}$ \quad
Jian Li$^{2}$ \quad
Di Wang$^{1,*}$\\[0.5em]
$^{1}$King Abdullah University of Science and Technology\\
$^{2}$Stony Brook University\\[0.3em]
\texttt{hlshen19@gmail.com, shaopeng.fu@kaust.edu.sa}\\
\texttt{qinbo.zhang@stonybrook.edu, jian.li.3@stonybrook.edu}\\
\texttt{di.wang@kaust.edu.sa}
}

\newcommand{\method}{\textsc{FOPL}}
\newcommand{\methodfull}{Final-Output Preference Learning}

\DeclareMathSizes{7.5}{7}{5}{5}

\newcommand{\tableformat}{%
    \centering\fontsize{7.5}{9}\selectfont
    \setlength{\tabcolsep}{3.5pt}%
    \renewcommand{\arraystretch}{1.10}%
}

\makeatletter
\newcommand{\tablecaption}[2]{%
    \begingroup
    \let\table@makecaption\@makecaption
    \long\def\@makecaption##1##2{%
        \table@makecaption{\fontsize{9}{10.5}\selectfont ##1}%
                         {\fontsize{9}{10.5}\selectfont ##2}%
    }%
    \caption{#1}\label{#2}%
    \endgroup
}
\makeatother

\tcbset{
  foplbox/.style={
    enhanced,
    breakable,
    colback=black!1,
    colframe=black!55,
    colbacktitle=black!16,
    coltitle=black,
    fonttitle=\bfseries\small,
    fontupper=\small,
    boxrule=0.6pt,
    arc=1.5mm,
    left=8pt,
    right=8pt,
    top=7pt,
    bottom=7pt,
    before skip=8pt,
    after skip=10pt,
    parbox=false,
  },
  foplprompt/.style={foplbox},
  foplexample/.style={foplbox},
}

\iclrfinalcopy
\begin{document}
\maketitle

\fancyhead{}
\fancyhead[L]{Preprint}
\thispagestyle{fancy}

\begingroup
\renewcommand{\thefootnote}{\fnsymbol{footnote}}
\footnotetext[1]{Corresponding author.}
\footnotetext[2]{This work was done during an internship at
King Abdullah University of Science and Technology.}
\endgroup

\begin{abstract}
Many large language models (LLMs) use chain-of-thought (CoT) to solve complex problems, but hide their CoT traces to prevent reasoning capabilities from being distilled.
While recent studies have shown that hidden CoTs can be reconstructed, existing methods typically require training a separate attacker for each target LLM.
This paper studies model-agnostic CoT reconstruction, aiming to reconstruct hidden CoT traces from arbitrary target LLMs using only input prompts and visible final responses.
We propose \methodfull{} (\method{}), which trains a single reconstruction model that generalizes to unseen LLMs without target-specific adaptation.
To achieve this, the reconstruction model must learn which reasoning patterns match a given final response without knowing the target LLM's identity.
We address this using preference optimization, training the reconstruction model to prefer native CoTs over synthetic alternatives for the same question and final response.
By learning these preferences from LLMs with accessible CoTs, the reconstruction model learns to distinguish reasoning patterns based on visible responses.
Experiments show that reconstructed CoTs capture distinct reasoning patterns across LLM families, including a family unseen during training.
Moreover, students trained on reconstructed CoTs match or outperform those trained on native CoTs across several reasoning benchmarks. The code is available at \url{https://github.com/renmiamu/fopl}.

\end{abstract}

\section{Introduction}

Recent advances in large language models (LLMs) have been driven in part by chain-of-thought (CoT) reasoning~\citep{wei2022chain,kojima2022large,wang2022self}, which enables models to solve complex problems through intermediate reasoning steps~\citep{jaech2024openai,comanici2025gemini,guo2025deepseek}.
These CoT traces are also valuable for model distillation~\citep{hinton2015distilling}, as training smaller models on the CoTs of stronger LLMs can significantly improve their reasoning capabilities~\citep{shridhar2023distilling,hsieh2023distilling,guo2025deepseek}.
However, this also creates a risk of model stealing, where attackers distill the reasoning capabilities of proprietary LLMs.
To reduce this risk, providers of frontier LLMs, such as GPT, Claude, and Gemini, typically hide their native CoTs and expose only final responses or brief reasoning summaries.

Recent studies have shown that hidden CoTs can still be extracted through API-based attacks.
\citet{panfilov2026stealing} prompt compatible LLMs to transcribe hidden CoTs from encrypted reasoning blocks, while \citet{luo2026capable} use custom tool calls to expose intermediate reasoning.
In contrast, \citet{zhang2026steal} propose to use supervised fine-tuning (SFT) on native CoTs from a surrogate LLM to train an attacker that reconstructs hidden CoTs from visible responses.
Once trained, this attacker can reconstruct CoTs without relying on specific API features or extraction prompts.
However, training on a single surrogate LLM may limit its ability to capture the distinct reasoning patterns of different target LLMs.
This raises a key question: \emph{Can a single attacker reconstruct target-specific reasoning patterns from the visible final responses of arbitrary LLMs?}

\begin{figure}[t]
    \centering
    \includegraphics[
        width=0.8\linewidth
    ]{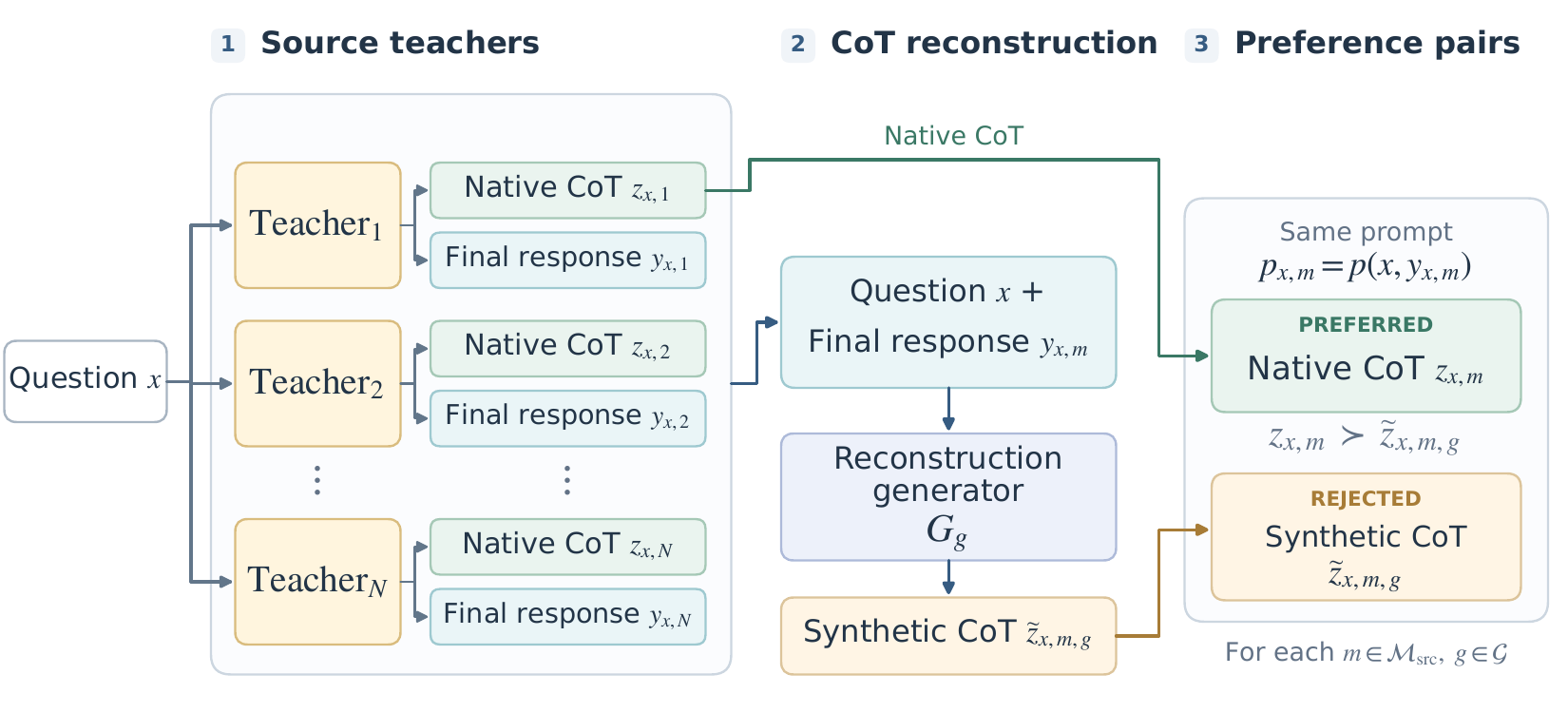}
    \caption{
    Preference-pair construction for \method{}.
    For each question, multiple source LLMs provide native CoTs and visible final responses.
    Reconstruction generators then produce synthetic CoTs conditioned on each source LLM's question and final response.
    We pair each native CoT (preferred) with synthetic alternatives (rejected) under the same question-response prompt, enabling DPO to train a single reconstruction model to distinguish reasoning patterns across different LLMs.
    }
    \label{fig:pipeline}
\end{figure}

In this paper, we propose \methodfull{} (\method{}), a model-agnostic CoT reconstruction method that trains a single attacker to reconstruct hidden CoTs from arbitrary target LLMs.
We collect native CoTs and final responses from multiple LLMs for the same questions.
A straightforward approach is to extend the SFT-based reconstruction method to this multi-LLM dataset.
However, SFT does not explicitly teach the attacker to distinguish which reasoning patterns best match each final response.
To address this challenge, we use Direct Preference Optimization (DPO)~\citep{rafailov2023direct} to train the attacker to distinguish reasoning patterns based on differences in visible responses.
As shown in Figure~\ref{fig:pipeline}, we construct preference pairs between native CoTs and synthetic alternatives generated by other LLMs, conditioned on the same question and final response.
By learning to prefer native CoTs over synthetic alternatives, the attacker learns to adapt its CoT reconstruction to the reasoning patterns reflected in each final response.

Empirically, we train a 4B attacker using FOPL with CoTs and final responses from four open-source surrogate LLMs.
We evaluate its effectiveness from two perspectives:
(1)~whether reconstructed CoTs resemble the target LLM's native reasoning traces,
and (2)~whether they provide effective supervision for downstream reasoning distillation.
For reconstruction quality, LLM-as-a-judge evaluations show that reconstructions conditioned on each target LLM's own final responses more closely resemble its native CoTs than those conditioned on other LLMs' responses.
This holds for both in-distribution and held-out LLM families.
For downstream distillation, we use a 1.5B student as a test case and evaluate its performance across mathematical reasoning, scientific reasoning, and safety benchmarks.
FOPL consistently outperforms existing CoT reconstruction baselines and often achieves better student performance than native-CoT distillation.
When applied to the proprietary GPT-5.6-Luna, FOPL improves student accuracy over final-response-only distillation by 8.93 and 6.40 percentage points on GSM8K and MATH-500, respectively.
Together, these results show that a single attacker can reconstruct CoTs across different LLMs without retraining, enabling effective reasoning distillation even when their native CoTs are hidden.

\section{Related Work}

\noindent\textbf{Model distillation.}
Knowledge distillation transfers capabilities from a larger teacher
to a smaller student~\citep{hinton2015distilling}, with sequence-level
methods using teacher-generated outputs as training
targets~\citep{kim2016sequence}.
Reasoning distillation extends this approach with intermediate
explanations and chain-of-thought (CoT), enabling smaller models
to learn reasoning beyond answer
prediction~\citep{magister2023teaching,shridhar2023distilling,
ho2023large}.
Distilling Step-by-Step uses rationales as auxiliary
supervision~\citep{hsieh2023distilling}, while Symbolic
Chain-of-Thought Distillation trains students on sampled reasoning
chains~\citep{li2023symbolic}.
Orca and DeepSeek-R1 further demonstrate the value of
reasoning demonstrations~\citep{mukherjee2023orca,guo2025deepseek}.

Subsequent work examines reasoning granularity and teacher
selection~\citep{chen2025unveiling}, and improves supervision
through example difficulty and diversity~\citep{li2025naturalthoughts}
and emphasis on key reasoning steps~\citep{dai2025capture}.
CoTD-PO further uses preference-based scoring to optimize
student-generated trajectories and reduce teacher--student
distribution mismatch~\citep{niu2025cotd}.
These methods improve how students learn from reasoning data,
while teachers that conceal their native CoTs introduce the
additional challenge of obtaining such supervision.

\medskip
\noindent\textbf{Reasoning trace extraction.}
Recent methods obtain hidden reasoning through target-model
interactions.
Reasoning Exposure Prompting elicits visible reasoning using
shadow-model demonstrations in code-like
formats~\citep{lu2026hidden}.
\citet{panfilov2026stealing} expose traces by passing encrypted
reasoning blocks to compatible models within the same provider's
ecosystem.
EchoCoT exploits reasoning replay through iterative tool
interactions~\citep{qu2026echocot}, while Forced-Reasoning
elicits intermediate reasoning through a custom tool
channel~\citep{luo2026capable}.
These methods depend on the target's elicitation behavior or
API capabilities.

Another line of research constructs reasoning from known outputs.
REverse-Engineered Reasoning synthesizes trajectories from
high-quality outputs for open-ended
generation~\citep{wang2026reverse}.
Studies of answer-conditioned generation show that instructions
to rationalize toward given answers can impair
distillation~\citep{lee2026answer}.
Structural Skeleton-guided Reasoning addresses answer anchoring
by guiding trace generation with a functional
skeleton~\citep{peng2026measuring}.
Trace Inversion~\citep{zhang2026steal} trains a reconstruction
model on native traces from a surrogate LLM, then generates traces
from target-model questions, final answers, and optional summaries
for student distillation.
\method{} further learns response-dependent reasoning patterns
across multiple surrogate LLMs.
By comparing native traces with synthetic alternatives under the
same question and final response, preference optimization trains
a single reconstruction model that transfers to unseen model
families without model identities or target-specific adaptation.

\section{Methodology}
\label{sec:method}

\subsection{Problem Setup}
\label{sec:problem-formulation}
\label{sec:online-stage}

Let $f$ denote a proprietary reasoning LLM that generates a chain-of-thought (CoT) trace $z$ followed by a final response $y$ for an input prompt $x$, i.e., $(z,y) \sim f(\cdot \mid x)$.
While the final response $y$ is visible to users, the native CoT $z$ remains hidden.
Our goal is to train a CoT reconstruction model $R_\theta$ that recovers the hidden CoT from the observable input-response pair $(x,y)$ as
\begin{align}
    \hat{z} \sim R_\theta(\cdot \mid x,y),
    \label{eq:cot_reconstruction}
\end{align}
where the reconstructed CoT $\hat{z}$ should closely resemble the native CoT $z$ in its reasoning patterns.

To train the CoT reconstruction model $R_\theta$, a straightforward approach is Trace-Inversion~\citep{zhang2026steal}, which trains $R_\theta$ using native CoTs from a surrogate LLM.
Specifically, given a CoT-accessible surrogate LLM $T$ and a prompt set $\mathcal{X}_s$, Trace-Inversion constructs a training dataset $\mathcal{D}_s = \{(x,y,z) \mid x \in \mathcal{X}_s, (z,y) \sim T(\cdot \mid x)\}$.
The reconstruction model is then trained to predict the native CoT $z$ from the corresponding prompt $x$ and final response $y$ by minimizing the following negative log-likelihood
\begin{align}
    \mathcal{L}_{\mathrm{SFT}}(\theta)
    = -\mathbb{E}_{(x,y,z)\sim\mathcal{D}_s}
    \left[\log R_\theta(z \mid x,y)\right].
    \label{eq:trace_inversion_sft}
\end{align}
However, training on a single surrogate LLM may limit generalization to targets with different reasoning patterns.
A natural extension is to collect native CoTs and final responses from multiple surrogate LLMs for the same questions.
Since different LLMs may follow different reasoning paths, the reconstruction model must learn which patterns match each final response.
Simply training on these examples does not explicitly teach the model to distinguish a target LLM's native reasoning patterns from other plausible alternatives.

\subsection{Final-Output Preference Learning}
\label{sec:offline-stage}

To explicitly teach the reconstruction model to distinguish native reasoning patterns from plausible alternatives, we propose \methodfull{} (\method{}), which learns these distinctions through preference optimization.
As illustrated in Figure~\ref{fig:pipeline}, the key idea of \method{} is to teach the reconstruction model to distinguish reasoning patterns by comparing native CoTs with synthetic alternatives conditioned on the same question and final response.
We use Direct Preference Optimization (DPO)~\citep{rafailov2023direct} to learn these preferences across multiple surrogate LLMs, enabling a single attacker to reconstruct CoTs that match different target LLMs' responses.
We now describe the proposed \method{} in detailed.

\noindent\textbf{Raw data collection from surrogate LLMs.}
Extending the single-surrogate setting in
Section~\ref{sec:problem-formulation}, we use $M$ CoT-accessible
surrogate LLMs, denoted by $T_m$, $m=1,\ldots,M$, and a shared
prompt set $\mathcal{X}_s$. For each prompt $x\in\mathcal{X}_s$,
we collect a native CoT and its corresponding visible final response
from each surrogate:
\begin{equation}
    (z_{x,m},y_{x,m})
    \sim T_m(\cdot\mid x).
\end{equation}
Here, $z_{x,m}$ and $y_{x,m}$ denote the native CoT and final response
produced by $T_m$, respectively. Following the notation of
$\mathcal{D}_s$ in Section~\ref{sec:problem-formulation}, the dataset
collected from each surrogate is
\begin{equation}
    \mathcal{D}_s^{(m)}
    =
    \left\{
        (x,y_{x,m},z_{x,m})
        \,\middle|\,
        x\in\mathcal{X}_s
    \right\}.
\end{equation}
All surrogates receive the same prompts, while their native CoTs
and final responses may differ. These datasets provide the observed
associations between final responses and native reasoning patterns
used to construct preference supervision.

\noindent\textbf{Preference data construction and model training.}
For each record in $\mathcal{D}_s^{(m)}$, we construct a reconstruction
prompt $p_{x,m}=p(x,y_{x,m})$ containing the question and visible final
response, without the surrogate's identity or native CoT. The full
template is provided in Appendix~\ref{app:reconstruction-prompt}.
We use $K$ reasoning models, denoted by $G_g$, $g=1,\ldots,K$,
to generate synthetic alternatives:
\begin{equation}
    \widetilde z_{x,m,g}
    \sim G_g(\cdot\mid p_{x,m}).
\end{equation}
The surrogate and generator pools may overlap. Even when $T_m$ and
$G_g$ are the same model, the synthetic trace is generated separately
from the response, without access to the native trace.

We pair each native trace $z_{x,m}$ with each synthetic alternative
$\widetilde z_{x,m,g}$ under the shared prompt $p_{x,m}$, treating the
native trace as preferred because it is the observed reasoning output
that accompanied $y_{x,m}$. We retain only prompts for which all required
native traces, final responses, and synthetic alternatives are
available. Let $\mathcal{X}_{\mathrm{pref}}\subseteq\mathcal{X}_s$
denote this retained subset. The preference dataset is
\begin{equation}
\begin{aligned}
    \mathcal{D}_{\mathrm{pref}}
    =
    \Bigl\{
        &(p_{x,m},z_{x,m},\widetilde z_{x,m,g})
        \,\Bigm|\,
        x\in\mathcal{X}_{\mathrm{pref}},
        m=1,\ldots,M,\quad g=1,\ldots,K
    \Bigr\}.
\end{aligned}
\end{equation}
With $N=|\mathcal{X}_{\mathrm{pref}}|$ retained prompts, this
construction yields $NMK$ preference pairs. Each comparison fixes
the question and final response, encouraging the reconstruction
model to distinguish reasoning patterns associated with the same
observed response.

We initialize $R_\theta$ from a frozen reference model $R_0$ and
optimize the standard DPO objective~\citep{rafailov2023direct}:
\begin{equation}
\begin{aligned}
    \mathcal{L}_{\mathrm{DPO}}(\theta)
    =
    -\mathbb{E}_{(p,z^+,z^-)\sim\mathcal{D}_{\mathrm{pref}}}
    \Biggl[
    \log\sigma\Biggl(
        \beta\Biggl[
        &\log\frac{R_\theta(z^+\mid p)}
                       {R_0(z^+\mid p)}
        -
        \log\frac{R_\theta(z^-\mid p)}
                       {R_0(z^-\mid p)}
        \Biggr]
    \Biggr)
    \Biggr],
\end{aligned}
\end{equation}
where each preference triple uses
$p=p_{x,m}$, $z^+=z_{x,m}$, and
$z^-=\widetilde z_{x,m,g}$.
Here, $\sigma$ is the logistic function, $\beta>0$ is the DPO
hyperparameter, and $R_0$ is the frozen reference model.
Conditioning on $p(x,y)$ implements the input-response conditioning
$R_\theta(\cdot\mid x,y)$ defined in
Section~\ref{sec:problem-formulation}; likelihoods are computed
only over CoT tokens.

We pool preference pairs from all surrogates to train one reconstruction
model, without supplying model identities as inputs. After training,
we freeze $R_{\theta^\star}$ and reconstruct CoTs for a target LLM
using only its input prompt and visible final response.

\section{Experiments}
\label{sec:experiments}

\subsection{Experiment Setup}
\label{sec:experimental-setup}

We evaluate whether reconstructed CoTs resemble the target model's native
traces and whether they provide useful supervision for student distillation.
These experiments assess trace similarity and downstream utility separately.
Higher student accuracy alone does not establish recovery of the target
model's original reasoning.

\noindent\textbf{Models.}
We use Qwen3-4B as the reconstruction model and
Qwen2.5-1.5B-Base~\citep{qwen2025qwen25technicalreport} as the
downstream student. DeepSeek, Kimi, GLM, and Qwen provide native CoTs
and visible final responses for preference training. We evaluate the
trained reconstruction model on these four model families and MiMo,
without supplying model identities or adapting the model to each
family. Downstream distillation uses responses from DeepSeek, Kimi,
GLM, and MiMo.

Many closed-source reasoning models do not expose their native CoTs,
making it difficult to compare reconstructed traces with the originals.
We therefore use MiMo, an open model family excluded from preference
training, as a controlled test of transfer to an unseen teacher.
When reconstructing from MiMo, \method{} receives only its questions
and visible final responses, as it would for a teacher whose CoTs are
hidden. We reserve MiMo's accessible native CoTs for measuring trace
similarity and for the native-CoT student reference. This setting tests
whether the reconstruction policy generalizes beyond the model families
used in preference training.

\noindent\textbf{CoT reconstruction quality.}
We use MATH-500~\citep{lightman2024let} to provide a common set of
mathematical problems for comparing reconstructed and native traces.
This experiment tests whether conditioning on a model's own response
produces a reconstruction more similar to its native CoT than conditioning
on another model's response to the same problem.

For each question $x$ and reference family $m$, we use the native trace
$z_{x,m}$ as the reference and generate
\[
    \widehat z_{x,m}
    \sim
    R_{\theta^\star}(\cdot\mid p(x,y_{x,m})).
\]
We compare this reconstruction with $\widehat z_{x,m'}$, generated from
the same question but another family's response $y_{x,m'}$, where
$m'\neq m$. A comparison is retained only when the normalized final
answers in the two responses agree. This controls for answer differences
while testing whether the visible response guides reconstruction toward
the reference model's reasoning patterns.

The judge receives the problem, the native reference, and two anonymous
reconstructions. It selects the more similar reconstruction or a tie,
considering solution strategy, decomposition, step order, verification,
explanatory emphasis, phrasing, and notation. Candidate order is
deterministically randomized, and model identities and conditioning
responses are hidden from the judge. We report the tie-aware preference
rate $(W+0.5T)/(W+L+T)$, where $W$, $L$, and $T$ are wins, losses, and
ties for the candidate being evaluated. Gemini 3.8 Flash is the primary
judge. GPT-5.6-Luna provides an independent evaluation in the
judge-robustness ablation (Section~\ref{sec:ablations}). The complete
judge prompt is provided in Appendix~\ref{app:judge-prompt}.

\noindent\textbf{Model distillation.}
We evaluate whether the reconstructed CoTs improve the reasoning
capabilities of a fixed student. For responses from DeepSeek, Kimi, GLM,
and MiMo, we train separate students using \method{} reconstructions,
synthetic CoTs generated by DeepSeek, Kimi, or GLM, and the response
model's native CoTs as a reference condition. All reconstructed and
synthetic traces are generated from the same question and visible
response. Within each response group, we also fix the question--response
records, student initialization, training configuration, and evaluation
settings. Only the intermediate CoT changes. MiMo's native traces are
used for evaluation and the student reference condition, but never to
train or condition a reconstruction model. We also compare against Trace Inversion
\citep{zhang2026steal}, an SFT-based CoT reconstruction baseline,
under the same downstream training and evaluation settings
(Table~\ref{tab:ti-baseline}).

We evaluate the trained students on GSM8K~\citep{cobbe2021training} and
MATH-500 to measure mathematical reasoning at grade-school and competition
levels, respectively. GPQA~\citep{rein2023gpqa} measures performance on
graduate-level science questions, while
OpenBookQA~\citep{mihaylov2018can} tests the application of elementary
science facts and background knowledge. These benchmarks measure the
utility of the traces as training supervision through the resulting
student's accuracy. We generate one response per question and report
single-run accuracy.

We additionally use harmful requests from
AdvBench~\citep{zou2023universal} to assess student response safety.
Llama Guard~\citep{grattafiori2024llama} classifies each generated response,
and we report the percentage classified as safe, with higher values
indicating better safety. Benchmark descriptions and evaluation generation
limits are provided in Appendix~\ref{app:evaluation-benchmarks}.

\noindent\textbf{Training details.}
Preference-training questions are sampled from
HARDTESTS~\citep{he2025hardtests}, and student-training records are
constructed from OpenR1-Math-220k. We train both the reconstruction model
and downstream student for three epochs with a learning rate of
$1\times10^{-5}$, and set $\beta=0.1$ for DPO. Data construction and
full training configurations are provided in
Appendix~\ref{app:training-details}.

\subsection{Main Results}
\label{sec:main-results}

\noindent {\bf Cross-family similarity.}
Table~\ref{tab:similarity} tests whether changing only the final-response
source changes the reconstruction in a source-consistent way while the
question and normalized answer are controlled. Each row fixes the native
reference family, and each column specifies the family whose final response
conditions \method{}.

\begin{table}[t]
\tablecaption{
Cross-family CoT similarity on MATH-500, judged by Gemini 3.8 Flash.
Rows indicate native-reference families; columns indicate the response
sources used by \method{}.
Entries are tie-aware preference rates (\%).
Bold diagonal entries aggregate comparisons against all four alternative
sources; off-diagonal entries report comparisons against the corresponding
column source.
}{tab:similarity}
\tableformat
\begin{tabular}{@{}lccccc@{}}
\toprule
Native reference
& \shortstack{DeepSeek\\response}
& \shortstack{Kimi\\response}
& \shortstack{GLM\\response}
& \shortstack{Qwen\\response}
& \shortstack{MiMo\\response} \\
\midrule
DeepSeek   & \textbf{57.46} & 45.85 & 39.85 & 40.98 & 43.48 \\
Kimi       & 32.41 & \textbf{62.98} & 35.52 & 42.46 & 37.69 \\
GLM        & 41.23 & 43.23 & \textbf{57.02} & 44.60 & 42.86 \\
Qwen       & 34.84 & 37.34 & 32.41 & \textbf{65.39} & 33.83 \\
\midrule
MiMo (OOD) & 36.90 & 43.98 & 36.31 & 40.73 & \textbf{60.51} \\
\bottomrule
\end{tabular}
\end{table}

Table~\ref{tab:similarity} shows a consistent diagonal advantage: for every
training family, reconstructions conditioned on that family's own final
responses are judged more similar to its native traces than those conditioned
on other families' responses. The matched-source tie-aware preference rates
range from approximately 57\% to 65\%, with the strongest effects for
Qwen and Kimi, while every mismatched-source comparison remains below 50\%.
Because the reconstruction policy and question are fixed across conditions,
this pattern suggests that final responses retain source-dependent
signals---such as solution strategy, explanatory emphasis, notation, or
phrasing---that guide reconstruction beyond final-answer agreement. This
interpretation is also consistent with the ablation results in
Section~\ref{sec:ablations}.

The diagonal advantage extends to MiMo, which is excluded from preference
training: its matched-source reconstruction achieves a tie-aware preference rate of
approximately 61\%. This result is less consistent with a purely memorized
mapping between the four training families and their typical styles, and
instead suggests that the policy can generalize to observable cues from an
unseen family. Nevertheless, the experiment does not isolate the responsible
cues or demonstrate the recovery of the model's latent reasoning process. We
therefore interpret the result as evidence of source-compatible reconstruction
from information present in the final response.

\noindent {\bf Downstream distillation.}
The goal of \method{} is to learn the source teacher's observable output
preferences and use them to construct student supervision when the teacher's
native CoT is unavailable. Table~\ref{tab:downstream} compares \method{} with
three synthetic-CoT baselines that reconstruct intermediate reasoning from
the same question and visible final response, without access to the source
teacher's native CoT. The native-CoT row is included separately as an oracle
reference representing the source teacher's original reasoning output. All
conditions within each response-source group differ only in the intermediate
reasoning trace used for student supervision.

\begin{table*}[t]
\tablecaption{
Downstream distillation into Qwen2.5-1.5B-Base. Within each response-source
group, the questions, visible final responses, student initialization, and
training configuration are fixed; only the intermediate CoT changes.
``Syn'' denotes a CoT synthesized from the question and visible final
response. Each native-CoT row is a reference for the source teacher's
original output preferences. ``Avg. $\Delta$'' reports the mean absolute
improvement over the base model across the five downstream tasks, in
percentage points. Bold and underlining mark the best and second-best
completed response-only conditions within each source and metric,
respectively; tied second-best results are both underlined. Native-CoT rows
are excluded from this ranking. MiMo is excluded from preference training.
Scores are percentages (higher is better).
}{tab:downstream}
\tableformat
\begin{tabular}{@{}llcccccc@{}}
\toprule
Response source
& Student supervision
& \shortstack{GSM8K\\Acc.}
& \shortstack{MATH-500\\Acc.}
& \shortstack{GPQA\\Acc.}
& \shortstack{OpenBookQA\\Acc.}
& \shortstack{AdvBench\\Safety}
& \shortstack{Avg. $\Delta$\\vs. Base} \\
\midrule
-- & Base model
& 27.22 & 15.00 & 3.54 & 0.20 & 6.92 & -- \\
\midrule

\multirow{5}{*}{DeepSeek}
& \method{} (Ours)
& \textbf{57.85} & \textbf{33.20} & \textbf{19.19}
& \textbf{50.30} & \underline{9.07} & +23.35 \\
& DeepSeek syn
& \underline{45.94} & \underline{24.80} & 15.66
& \underline{47.30} & \textbf{12.88} & +18.74 \\
& Kimi syn
& 33.66 & 20.80 & \underline{18.69}
& 34.60 & 5.58 & +12.09 \\
& GLM syn
& 37.07 & 18.40 & 18.18
& 40.30 & 6.35 & +13.48 \\
\cmidrule(lr){2-8}
& Native CoT (Ref)
& 49.28 & 23.00 & 16.67
& 56.90 & 15.96 & +21.79 \\
\midrule

\multirow{5}{*}{Kimi}
& \method{} (Ours)
& \textbf{53.30} & \textbf{27.60} & \textbf{17.17}
& \underline{43.00} & \textbf{12.12} & +20.06 \\
& DeepSeek syn
& \underline{50.64} & \underline{26.60} & \underline{16.16}
& 37.00 & \underline{7.12} & +16.93 \\
& Kimi syn
& 45.56 & 25.40 & \underline{16.16}
& 42.20 & 6.15 & +16.52 \\
& GLM syn
& 47.69 & 15.80 & 15.66
& \textbf{43.60} & 5.77 & +15.13 \\
\cmidrule(lr){2-8}
& Native CoT (Ref)
& 56.94 & 27.20 & 14.14
& 37.40 & 20.00 & +20.56 \\
\midrule

\multirow{5}{*}{GLM}
& \method{} (Ours)
& \textbf{52.62} & \textbf{29.40} & \textbf{18.69}
& \textbf{39.20} & \underline{11.73} & +19.75 \\
& DeepSeek syn
& \underline{48.52} & \underline{26.40} & 12.12
& 34.80 & \textbf{12.50} & +16.29 \\
& Kimi syn
& 42.53 & 25.60 & \underline{16.16}
& 37.40 & 6.35 & +15.03 \\
& GLM syn
& 46.40 & 22.40 & 15.15
& \underline{39.00} & 10.38 & +16.09 \\
\cmidrule(lr){2-8}
& Native CoT (Ref)
& 48.37 & 27.40 & 16.16
& 43.60 & 21.92 & +20.91 \\
\midrule

\multirow{5}{*}{MiMo (OOD)}
& \method{} (Ours)
& \textbf{49.58} & \textbf{22.60}
& \textbf{19.70} & \textbf{35.60}
& \textbf{15.77} & +18.07 \\
& DeepSeek syn
& 33.97 & 15.40
& 14.65 & 13.00
& 13.85 & +7.60 \\
& Kimi syn
& 44.88 & \underline{19.20}
& 13.64 & 19.00
& 13.87 & +11.54 \\
& GLM syn
& \underline{47.08} & 12.40
& \underline{16.16} & \underline{27.60}
& \underline{14.26} & +12.92 \\
\cmidrule(lr){2-8}
& Native CoT (Ref)
& 53.98 & 18.60
& 11.62 & 25.00
& 15.77 & +14.42 \\
\bottomrule
\end{tabular}
\end{table*}

First, \method{} consistently provides stronger student supervision than direct prompted synthesis.
Across all four response sources, \method{} achieves the largest average improvement over the base model among the reconstruction conditions, with Avg.~$\Delta$ ranging from +18.07 to +23.35 points. The advantage is particularly consistent on the reasoning-oriented benchmarks: for DeepSeek, Kimi, and GLM responses, \method{} achieves the best performance among methods that reconstruct CoTs from the same question and visible final response on GSM8K, MATH-500, and GPQA, rather than benefiting from only one favorable task or response source. The results therefore suggest that learning source-conditioned reasoning preferences provides a more reliable supervision signal than simply asking another strong model to produce an answer-consistent CoT.

Second, reconstructed supervision can reach or exceed the downstream utility of the source teacher's native CoT, but the advantage is task dependent.
Across the four response sources, \method{} outperforms native-CoT supervision
in 10 of the 12 comparisons on GSM8K, MATH-500, and GPQA. Native CoT performs
better in three of the four AdvBench comparisons, showing that the advantage
does not extend uniformly to safety. The native trace serves as a reference
for the teacher's original reasoning preferences rather than an upper bound
on student performance.

Third, the MiMo results show that this supervision advantage transfers to a teacher family excluded from preference training.
Although MiMo is never used to construct the offline preference data, \method{} still obtains an Avg.~$\Delta$ of +18.07 points, compared with +7.60 to +12.92 for the three prompted synthetic baselines. The advantage is especially clear on MATH-500 and GPQA, where reconstructed supervision also exceeds the MiMo native-CoT reference. Combined with the source-matching result in Table~\ref{tab:similarity}, this provides evidence that \method{} is not simply memorizing the reasoning styles of the training families. Instead, it learns response-conditioned reconstruction patterns that remain useful for an unseen response source, directly supporting the intended setting in which the target teacher provides only visible final responses.

\noindent\textbf{Comparison with Trace Inversion.}
We compare \method{} with Trace Inversion (TI)
\citep{zhang2026steal} to evaluate its downstream utility against
an existing CoT reconstruction method. We reproduce TI in the
no-summary setting using the same OpenThoughts-114k dataset
\citep{guha2026openthoughts} as the original work, which provides
DeepSeek-R1 reasoning traces \citep{guo2025deepseek}.
We replace the original TI backbone with Qwen3-4B, matching
the reconstruction backbone used by \method{}.
TI is trained with token-level cross-entropy to reconstruct
CoTs from questions and visible final responses.
Within each response-source group, both methods use the same
retained questions, final responses, downstream student
initialization, SFT configuration, and evaluation settings.

\begin{table*}[t]
\tablecaption{
Downstream distillation using CoTs reconstructed by \method{}
and Trace Inversion (TI). Both methods condition on the same
questions and visible final responses within each response-source
group. The student is Qwen2.5-1.5B-Base. Scores are percentages,
and higher values are better.
}{tab:ti-baseline}
\tableformat
\begin{tabular}{@{}llcccc@{}}
\toprule
Response source
& Reconstructor
& \shortstack{GSM8K\\Acc.}
& \shortstack{MATH-500\\Acc.}
& \shortstack{GPQA\\Acc.}
& \shortstack{OpenBookQA\\Acc.} \\
\midrule
\multirow{2}{*}{DeepSeek}
& \method{} & 57.85 & 33.20 & 19.19 & 50.30 \\
& TI (R1) & 28.51 & 17.00 & 8.59 & 15.00 \\
\midrule
\multirow{2}{*}{Kimi}
& \method{} & 53.30 & 27.60 & 17.17 & 43.00 \\
& TI (R1) & 38.06 & 16.40 & 8.08 & 19.00 \\
\midrule
\multirow{2}{*}{GLM}
& \method{} & 52.62 & 29.40 & 18.69 & 39.20 \\
& TI (R1) & 33.36 & 13.40 & 8.59 & 9.60 \\
\bottomrule
\end{tabular}
\end{table*}

Table~\ref{tab:ti-baseline} shows that \method{} outperforms TI
in all 12 comparisons across three response sources and four
benchmarks. The advantage holds for DeepSeek responses as well
as Kimi and GLM responses, showing that the learned reconstruction
policy provides effective supervision across different target
teachers. Together with the prompted-synthesis comparisons in
Table~\ref{tab:downstream}, these results demonstrate the
downstream utility of \method{} relative to both prompted and
SFT-based reconstruction baselines.

\subsection{Ablation Studies}
\label{sec:ablations}

\noindent {\bf SFT versus DPO reconstruction.}
To examine the effect of preference training, we train an SFT reconstruction
policy on the same multi-family native traces used as the chosen examples
in the DPO pairs. The reconstruction model, number of training examples,
and training hyperparameters are matched, while DPO additionally uses the
corresponding rejected traces. We compare the two reconstruction policies
in terms of downstream distillation and source matching.

For downstream distillation, both policies reconstruct traces from the same
DeepSeek questions and visible final responses. The resulting traces are
used to train the same Qwen2.5-1.5B-Base student under the same downstream
settings. Table~\ref{tab:revision-matched-sft} shows that DPO yields higher
student accuracy on all five benchmarks, with gains ranging from 2.42 to
16.70 percentage points. These results support the value of learning from
native-versus-synthetic preferences over SFT on native traces alone.

\begin{table}[htbp]
\tablecaption{Matched SFT and DPO reconstruction policies: downstream student
accuracy on five benchmarks (\%). Both policies use the
same native traces as their positive training examples, while DPO additionally
uses the corresponding rejected traces. The last row reports the difference in
percentage points.}{tab:revision-matched-sft}
\tableformat
\begin{tabular}{@{}lccccc@{}}
\toprule
Reconstructor & GSM8K & MATH-500 & GPQA & OpenBookQA & MBPP \\
\midrule
SFT & 48.67 & 19.60 & 13.13 & 33.60 & 29.80 \\
DPO & \textbf{57.85} & \textbf{33.20} & \textbf{19.19} & \textbf{50.30} & \textbf{32.22} \\
DPO $-$ SFT & +9.18 & +13.60 & +6.06 & +16.70 & +2.42 \\
\bottomrule
\end{tabular}
\end{table}

For source matching, we evaluate both policies on MATH-500 using DeepSeek,
Kimi, GLM, and the held-out MiMo family as the four response sources.
Gemini 3.8 Flash judges similarity to each family's native reference trace,
with the matching-source candidate compared against candidates conditioned
on the other three sources. Table~\ref{tab:revision-sft-matching} reports
the SFT preference rates and their differences from DPO under this
four-source comparison.

\begin{table}[t]
\tablecaption{MATH-500 source matching with the SFT reconstruction policy.
Entries are Gemini tie-aware preference rates (\%). Parentheses report
SFT $-$ DPO differences in percentage points under the same four-source
comparison, and bold marks the matching source. Each diagonal aggregates
three comparisons, while each off-diagonal
entry reports the candidate's rate against the same-source candidate.}
{tab:revision-sft-matching}
\tableformat
\centering
\setlength{\tabcolsep}{3pt}
\renewcommand{\arraystretch}{1.05}
\begin{tabular}{@{}lcccc@{}}
\toprule
Native reference & DeepSeek syn & Kimi syn & GLM syn & MiMo syn \\
\midrule
DeepSeek & \textbf{53.83} ($-$3.11) & 47.01 (+1.16) & 45.01 (+5.16) & 46.50 (+3.02) \\
Kimi     & 39.32 (+6.91) & \textbf{61.68} ($-$3.11) & 35.38 ($-$0.14) & 40.27 (+2.58) \\
GLM      & 42.00 (+0.77) & 48.49 (+5.26) & \textbf{54.88} ($-$2.68) & 44.88 (+2.02) \\
MiMo     & 44.00 (+7.10) & 48.62 (+4.64) & 45.39 (+9.08) & \textbf{54.00} ($-$6.93) \\
\bottomrule
\end{tabular}
\end{table}

DPO achieves higher matching-source preference rates for all four families,
exceeding SFT by 3.11, 3.11, 2.68, and 6.93 percentage points for DeepSeek,
Kimi, GLM, and MiMo, respectively. The largest gain occurs on MiMo, which
is excluded from preference training. Together with the downstream results,
this comparison shows that preference training improves both source
alignment and the utility of reconstructed supervision, including source
alignment for an unseen teacher family.

\noindent {\bf Robustness to the judge model.}
To determine whether the cross-family alignment results depend on a particular
evaluator, we compare the primary Gemini 3.8 Flash judgments with judgments
from GPT-5.6-Luna. Both judges evaluate the same candidate pairs using the same
similarity rubric and candidate-order randomization. For each native-reference
family, we report the matching-source rate, the average mismatched-source rate,
and their difference. We additionally report the percentage of shared valid
comparisons on which the two judges make the same decision.

\begin{table*}[t]
\tablecaption{
Robustness of cross-family alignment to the judge model. ``Matched'' is the
preference rate of the reconstruction conditioned on the reference family's
own final response. ``Mismatched'' averages the preference rates of
reconstructions conditioned on the other four response sources, and ``Gap''
is Matched minus Mismatched. Agreement is the percentage of shared valid
comparisons for which Gemini 3.8 Flash and GPT-5.6-Luna select the same
underlying candidate after their decisions are mapped back to the original
candidate identities. Ties are treated as a separate outcome. MiMo is held
out of preference training.
}{tab:judge-ablation}
\tableformat
\setlength{\tabcolsep}{3pt}
\begin{tabular}{@{}lccccccc@{}}
\toprule
& \multicolumn{3}{c}{Gemini 3.8 Flash}
& \multicolumn{3}{c}{GPT-5.6-Luna}
& \multirow{2}{*}{\shortstack{Judge\\agreement}} \\
\cmidrule(lr){2-4}
\cmidrule(lr){5-7}
Native reference
& Matched
& Mismatched
& Gap
& Matched
& Mismatched
& Gap
& \\
\midrule
DeepSeek
& 57.46 & 42.54 & 14.92
& 55.22 & 44.78 & 10.44
& 70.93 \\

Kimi
& 62.98 & 37.02 & 25.96
& 57.77 & 42.23 & 15.54
& 67.50 \\

GLM
& 57.02 & 42.98 & 14.04
& 54.98 & 45.02 & 9.96
& 69.51 \\

Qwen
& 65.39 & 34.61 & 30.79
& 56.67 & 43.33 & 13.33
& 69.35 \\

MiMo (OOD)
& 60.51 & 39.48 & 21.03
& 57.98 & 42.02 & 15.95
& 72.94 \\

\midrule
Average
& 60.67 & 39.33 & 21.35
& 56.52 & 43.48 & 13.05
& 70.05 \\
\bottomrule
\end{tabular}
\end{table*}

Table~\ref{tab:judge-ablation} shows that the matching-source pattern is robust
to the choice of evaluator. GPT-5.6-Luna reproduces the same qualitative result
as Gemini 3.8 Flash: for every native-reference family, the matching-source
reconstruction receives a higher preference rate than reconstructions
conditioned on mismatched response sources. Gemini produces a larger average
alignment gap, but the gap remains positive for every family under Luna,
indicating that the diagonal structure in Table~\ref{tab:similarity} is not
specific to one evaluator's preferences.

The two judges agree on 70.05\% of their shared valid comparisons, with
family-level agreement ranging from 67.50\% to 72.94\%. Importantly, both
judges preserve a positive matching-source advantage for the held-out MiMo
family, for which they obtain their highest agreement. Although individual
decisions differ across some comparisons, the aggregate source-matching
conclusion remains stable across the two evaluator families.

\noindent {\bf Final-response-only supervision.}
We remove the reconstructed reasoning trace and train the student only on the
visible DeepSeek final response. This control tests whether the gains of
\method{} can be explained solely by exposing the student to the teacher's
visible output. All conditions use the same question--response records,
student initialization, and SFT configuration.

\begin{figure}[t]
    \tablecaption{
    Final-response-only ablation on GSM8K and MATH-500. We compare
    supervision using the source model's native CoT, the visible final response
    alone, and the reasoning trace reconstructed by \method{}. All conditions
    use the same DeepSeek question--response records and student-training
    configuration.
    }{fig:final-only}

    \centering
    \begin{minipage}[t]{0.48\linewidth}
        \centering
        \includegraphics[width=\linewidth]{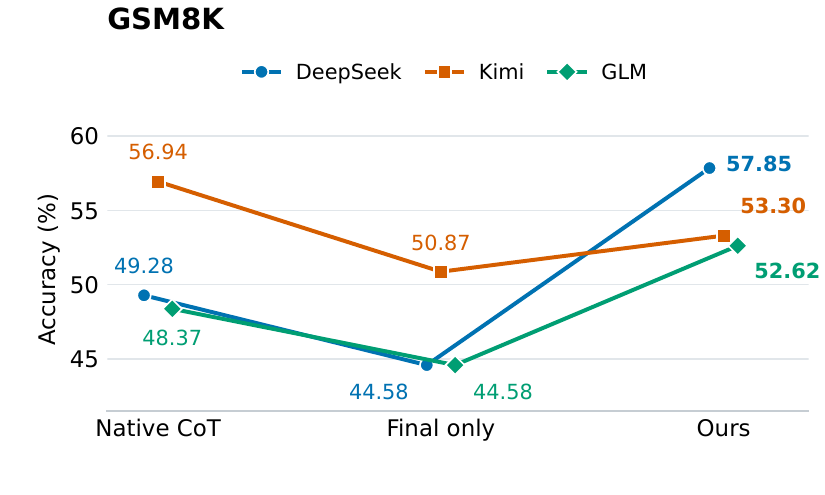}
    \end{minipage}
    \hfill
    \begin{minipage}[t]{0.48\linewidth}
        \centering
        \includegraphics[width=\linewidth]{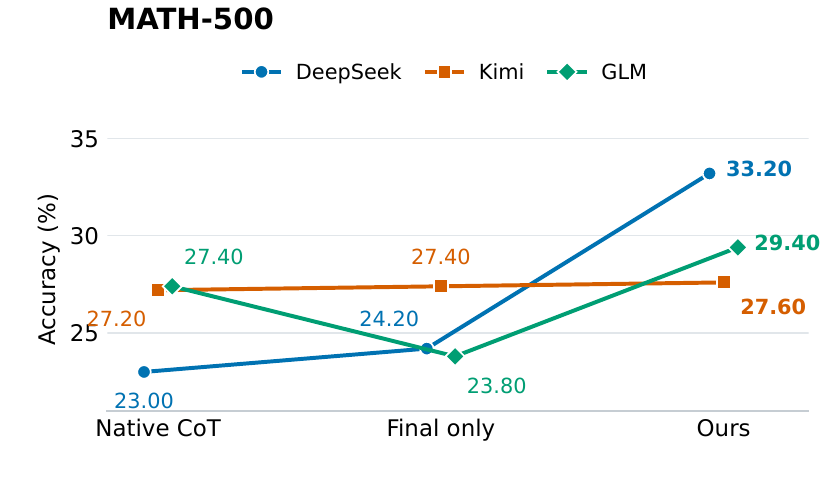}
    \end{minipage}
\end{figure}

As shown in Figure~\ref{fig:final-only}, final-response-only training
underperforms \method{} on both mathematics benchmarks. Although the visible
response contains the answer and teacher-specific presentation cues, it does
not provide the same explicit intermediate supervision as a reconstructed
trace. The advantage of \method{} therefore comes from converting information
in the final response into a structured reasoning demonstration, rather than
merely adding the teacher's visible output to the student-training data.

\noindent\textbf{Effect of downstream SFT data scale.}
We evaluate how the amount of reconstructed supervision affects
student performance. Keeping the trained \method{} reconstruction
policy fixed, we fine-tune the student on 1,250, 2,500, and 5,000
examples, corresponding to 25\%, 50\%, and 100\% of the same
DeepSeek response pool. Smaller subsets are nested within larger
ones, and all conditions use the same student initialization and
SFT hyperparameters.

\begin{table}[t]
\tablecaption{
Effect of downstream SFT data scale. All fine-tuned conditions
use traces reconstructed by the same fully trained \method{}
policy from the DeepSeek response pool and differ only in the
number of student-training examples. The base model is included
as a reference. Scores are percentages, and higher values are better.
}{tab:data-scale-ablation}
\tableformat
\begin{tabular}{@{}lccccc@{}}
\toprule
SFT data
& Training examples
& \shortstack{GSM8K\\Acc.}
& \shortstack{MATH-500\\Acc.}
& \shortstack{GPQA\\Acc.}
& \shortstack{OpenBookQA\\Acc.} \\
\midrule
Base model & --    & 27.22 & 15.00 & 3.54  & 0.20 \\
25\%       & 1,250 & 51.55 & 28.40 & 17.68 & 29.60 \\
50\%       & 2,500 & 54.06 & 27.80 & 20.20 & 38.80 \\
100\%      & 5,000 & 57.85 & 33.20 & 19.19 & 50.30 \\
\bottomrule
\end{tabular}
\end{table}

Table~\ref{tab:data-scale-ablation} shows that reconstructed
supervision improves student performance over the base model
on all four benchmarks even with only 1,250 training examples.
Using all 5,000 examples yields the highest accuracy on GSM8K,
MATH-500, and OpenBookQA, while GPQA peaks at 2,500 examples.
These results show that the reconstructed traces provide useful
supervision with limited downstream data and that increasing
their quantity generally improves student performance.

\section{Conclusion}

We introduce \method{}, a final-response-conditioned preference-learning
framework for constructing reasoning supervision when a target teacher's
native chain-of-thought is unavailable. During offline training, preference
pairs hold the question and visible final response fixed while contrasting
native traces with answer-consistent synthetic alternatives. This teaches a
single reconstruction policy to favor reasoning patterns associated with
source teachers' observable outputs. At transfer time, the policy generates
a trace using only the target question and final response, without requiring
the target teacher's native trace or identity. Experiments show that the
policy produces source-consistent traces for a held-out teacher family and
provides effective supervision for downstream students. Together, these
results show that visible final responses carry transferable signals of
teacher output preferences that can be turned into useful reasoning
supervision even when the target teacher's native CoT is inaccessible.

\clearpage

\subsection*{AI use statement}

We used generative AI models as experimental components to produce teacher
responses and synthetic reasoning traces and to evaluate trace similarity.
We also used an AI assistant to improve the clarity and readability of the
manuscript, including language polishing and correction of grammatical errors.
The authors reviewed the AI-assisted edits and take responsibility for the
final text.

\subsection*{Ethics statement}

Our method generates plausible reasoning traces from visible teacher
responses. These traces should not be interpreted as verbatim records of a
target model's internal reasoning. We evaluate their similarity to observable
teacher outputs and their utility as student supervision.

\subsection*{Reproducibility statement}

The main text describes the preference-pair construction, reconstruction
method, downstream comparisons, and evaluation protocol. Additional details
on the data and training configuration are provided in
Appendix~\ref{app:training-details}, and the similarity-judge prompt is
provided in Appendix~\ref{app:judge-prompt}.

\clearpage
\bibliography{iclr2027_conference}
\bibliographystyle{iclr2027_conference}

\clearpage
\appendix


\section{Data and Training Configuration}
\label{app:training-details}

\subsection{Preference Data}
\label{app:preference-data}

The preference source pool contains 500 programming problems sampled from
HARDTESTS~\citep{he2025hardtests}. Both the target-teacher pool
$\mathcal{M}_{\mathrm{train}}$ and the reconstruction-generator pool
$\mathcal{G}$ contain DeepSeek-V4-Pro, Kimi-K2.5, GLM-5.2, and
Qwen3-235B-A22B-Thinking-2507.

Following the construction in Section~\ref{sec:offline-stage}, each target
teacher $T_m$ produces a native trace $z_{x,m}$ and visible final response
$y_{x,m}$ for every source question $x$. We construct the response-conditioned
prompt
\begin{equation}
    p_{x,m}=p(x,y_{x,m})
\end{equation}
and use each reconstruction generator $G_g$ to produce
\begin{equation}
    \widetilde z_{x,m,g}
    \sim
    G_g(\,\cdot\mid p_{x,m}).
\end{equation}
Each target--generator combination therefore contributes the preference triple
\begin{equation}
    \left(
        p_{x,m},
        z_{x,m},
        \widetilde z_{x,m,g}
    \right),
\end{equation}
where the native trace is preferred and the reconstructed trace is rejected.

A problem is retained only when all four target teachers provide nonempty
native traces and final responses and all $4\times4$ target--generator
reconstructions are available. After this filtering, the exported preference
dataset contains \textbf{6,928 pairs from 433 distinct problem IDs}.

Each retained problem contributes
\begin{equation}
    \left|\mathcal{M}_{\mathrm{train}}\right|
    \times
    \left|\mathcal{G}\right|
    =
    4\times4
    =
    16
\end{equation}
preference pairs. Every target--generator combination therefore contributes
433 examples. The four same-family combinations contribute
\begin{equation}
    4\times433=1{,}732
\end{equation}
pairs, while the twelve cross-family combinations contribute
\begin{equation}
    12\times433=5{,}196
\end{equation}
pairs.

\subsection{Reconstruction Prompt}
\label{app:reconstruction-prompt}

We use the following prompt to reconstruct an intermediate reasoning trace from
a question and its reference answer. The reference answer is provided only as
guidance for determining the direction and organization of the reconstruction;
the generated trace is instructed not to reveal or imply access to the
reference answer.

\begin{tcolorbox}[foplprompt,title={Reconstruction Prompt}]
Below are a question and its reference answer. Develop a
\textbf{clear, step-by-step line of reasoning} that arrives at the
reference answer, expressed as a \textbf{first-person think-aloud
monologue}.

Use the reference answer \textbf{privately} to guide the direction
and organization of the reasoning. The resulting monologue
should nevertheless sound like an \textbf{authentic solution being
worked out in real time on a first encounter}, with \textbf{no mention
or suggestion that the reference answer was available in advance}.

\medskip
\noindent
\textbf{Question:} \texttt{\{question\}}

\noindent
\textbf{Reference Answer:} \texttt{\{reference\_answer\}}

\medskip
\noindent
\textbf{OUTPUT REQUIREMENTS:}

\begin{enumerate}
    \item Return \textbf{ONLY the first-person think-aloud monologue}.
    \item Write in the voice of a \textbf{focused person reasoning
    privately to themself}.
    \item \textbf{Never state or imply} that you previously saw or used the
    Reference Answer.
\end{enumerate}
\end{tcolorbox}

\subsection{Downstream Student Data}
\label{app:student-data}

Downstream question--response records are constructed from
OpenR1-Math-220k. The checked student-training
configuration caps the input at 5,000 records and does not reserve a held-out
SFT validation split.

For each downstream record $(x_i,y_i)$, the trained reconstruction policy
generates a reasoning trace using only the question $x_i$ and the visible final
response $y_i$. The reconstructed trace and final response are then serialized
into a single student target. Native teacher traces are required for offline
preference training and for an oracle reference condition when available; they
are not required to generate \method{} supervision for a target teacher.

Within each response-source block in Table~\ref{tab:downstream}, comparisons
use the same retained question IDs, exact final responses, preprocessing
pipeline, student initialization, SFT configuration, and evaluation decoding
settings.

\subsection{Training Configuration}
\label{app:training-config}

Table~\ref{tab:training-config} summarizes the checked repository
configuration for reconstruction-policy DPO and downstream student SFT.
During DPO, all reconstruction-policy parameters are updated, while a frozen
copy of the initialization is used as the reference policy.

\begin{table}[ht]
\tablecaption{
Configured reconstruction-DPO and student-SFT settings.
Executed-run overrides should be verified against the final training logs.
}{tab:training-config}
\tableformat
\fontsize{9}{11}\selectfont
\setlength{\tabcolsep}{6pt}
\renewcommand{\arraystretch}{1.15}
\begin{tabular}{@{}lcc@{}}
\toprule
Setting & Reconstruction DPO & Student SFT \\
\midrule
Epochs                       & 3          & 3          \\
Learning rate                & $10^{-5}$  & $10^{-5}$  \\
Schedule                     & Cosine     & Cosine     \\
Warmup ratio                 & 5\%        & 5\%        \\
Device batch size            & 1          & 1          \\
Gradient accumulation        & 8          & 8          \\
Maximum sequence length      & 16,384     & 32,768     \\
Maximum prompt length        & 8,192      & --         \\
DPO $\beta$                  & 0.1        & --         \\
Precision                    & bfloat16   & bfloat16   \\
Gradient checkpointing       & Enabled    & Enabled    \\
Seed                         & 42         & 42         \\
\bottomrule
\end{tabular}
\end{table}

\subsection{Evaluation Benchmarks and Generation Limits}
\label{app:evaluation-benchmarks}

GSM8K~\citep{cobbe2021training} evaluates multi-step reasoning on
grade-school mathematics word problems. MATH-500~\citep{lightman2024let}
contains 500 problems from the MATH benchmark and evaluates mathematical
problem solving. GPQA~\citep{rein2023gpqa} evaluates graduate-level
multiple-choice questions in biology, physics, and chemistry.
OpenBookQA~\citep{mihaylov2018can} evaluates multiple-choice elementary
science questions that require applying scientific facts and background
knowledge. AdvBench~\citep{zou2023universal} contains harmful requests and
is used here to assess the safety of generated responses.

For downstream student evaluation, the maximum number of generated tokens
per response is 8,192 for GSM8K and OpenBookQA, 32,768 for MATH-500, and
16,384 for GPQA. These are inference-time generation limits, separate from
the student-SFT maximum sequence length in
Table~\ref{tab:training-config}.

\section{Supplementary Closed-Source Teacher Transfer}
\label{app:closed-source-transfer}

As supplementary evidence, we evaluate whether \method{} can support
distillation from GPT-5.6-Luna, a closed-source teacher that exposes visible
final responses but no native reasoning traces. Let $y_{\mathrm{L}}$ denote a
visible Luna response. The frozen reconstruction policy generates
\[
    \widehat z_{\mathrm{L}}
    \sim
    R_{\theta^\star}(\cdot\mid p(x,y_{\mathrm{L}}))
\]
without access to a Luna-native reasoning trace or teacher identifier. We
compare a student trained on reconstructed targets
$h(\widehat z_{\mathrm{L}},y_{\mathrm{L}})$ with a student trained only on
the visible Luna response $y_{\mathrm{L}}$. Both conditions use the same
question IDs, final responses, student initialization, preprocessing, SFT
hyperparameters, and evaluation settings.

\begin{table}[t]
\tablecaption{
Closed-source teacher transfer from GPT-5.6-Luna to Qwen2.5-1.5B-Base.
GPT-5.6-Luna exposes only visible final responses and does not provide native
reasoning traces. The two fine-tuned conditions use the same questions, Luna
responses, student initialization, and training configuration. Scores are
percentages, and higher values are better.
}{tab:closed-source-transfer}
\tableformat
\begin{tabular}{@{}lcc@{}}
\toprule
Student supervision
& \shortstack{GSM8K\\Acc.}
& \shortstack{MATH-500\\Acc.} \\
\midrule
Base model
& 27.22 & 15.00 \\
Luna final response only
& 42.82 & 23.40 \\
\method{} reconstruction
& 51.75 & 29.80 \\
\bottomrule
\end{tabular}
\end{table}

Reconstructed supervision outperforms final-response-only training by
8.93 points on GSM8K and 6.40 points on MATH-500. Both conditions use the
same questions, Luna responses, student initialization, and training
configuration, so the comparison isolates the contribution of the
reconstructed trace. The gains suggest that making intermediate reasoning
explicit provides more useful supervision for this student than training on
the visible response alone. The reconstruction policy receives neither a
Luna family identifier nor a native Luna trace. Its performance therefore
provides supplementary evidence that the learned reconstruction patterns
transfer beyond the teacher families used for preference training.


\section{End-to-End Training Procedure}
\label{app:algorithm}

Algorithm~\ref{alg:fopl} summarizes the three stages of \method{}:
preference-data construction, reconstruction-policy optimization, and
downstream student-supervision generation.

Native traces are used as preferred responses during preference-data
construction and reconstruction-policy training. Once DPO training is
complete, the reconstruction policy requires only a question and its visible
final response.

\begin{algorithm}[t]
\caption{\method{} training and downstream supervision construction}
\label{alg:fopl}
\small
\begin{algorithmic}[1]

\Statex \textbf{Input:}
Question pool $\mathcal{Q}$;
teacher families $\mathcal{M}_{\mathrm{train}}$;
reconstruction generators $\mathcal{G}$;
initial reconstruction policy $R_0$;
downstream response data $\mathcal{D}_{\mathrm{resp}}$;
serialization function $h$.

\Statex \textbf{Output:}
Trained reconstruction policy $R_{\theta^\star}$ and student-supervision
dataset $\mathcal{D}_{\mathrm{SFT}}$.

\Statex
\Statex \textbf{Stage 1: Preference-data construction}

\State $\mathcal{D}_{\mathrm{pref}}\gets\varnothing$

\For{each question $x\in\mathcal{Q}$}
    \State $\mathcal{C}_x\gets\varnothing$
    \State $\mathrm{complete}\gets\mathrm{true}$

    \For{each teacher $m\in\mathcal{M}_{\mathrm{train}}$}
        \State Sample native output
        $(z_{x,m},y_{x,m})\sim T_m(\cdot\mid x)$

        \If{$z_{x,m}$ or $y_{x,m}$ is empty}
            \State $\mathrm{complete}\gets\mathrm{false}$
        \Else
            \State Construct shared prompt
            $p_{x,m}\gets p(x,y_{x,m})$

            \For{each generator $g\in\mathcal{G}$}
                \State Sample reconstruction
                $\widetilde z_{x,m,g}\sim
                G_g(\cdot\mid p_{x,m})$

                \If{$\widetilde z_{x,m,g}$ is empty}
                    \State $\mathrm{complete}\gets\mathrm{false}$
                \Else
                    \State Add
                    $(p_{x,m},z_{x,m},\widetilde z_{x,m,g})$
                    to $\mathcal{C}_x$
                \EndIf
            \EndFor
        \EndIf
    \EndFor

    \If{$\mathrm{complete}$ and
        $\lvert\mathcal{C}_x\rvert
        =
        \lvert\mathcal{M}_{\mathrm{train}}\rvert
        \lvert\mathcal{G}\rvert$}
        \State
        $\mathcal{D}_{\mathrm{pref}}
        \gets
        \mathcal{D}_{\mathrm{pref}}\cup\mathcal{C}_x$
    \EndIf
\EndFor

\Statex
\Statex \textbf{Stage 2: Reconstruction-policy optimization}

\State Initialize trainable policy $R_\theta\gets R_0$
\State Keep reference policy $R_0$ frozen

\For{each DPO training epoch}
    \For{each $(p,z^+,z^-)\in\mathcal{D}_{\mathrm{pref}}$}
        \State Compute implicit reward
        \[
            r_\theta(z;p)
            =
            \beta
            \log
            \frac{R_\theta(z\mid p)}
                 {R_0(z\mid p)}
        \]
        \State Update $\theta$ by minimizing
        \[
            -\log\sigma\!\left(
                r_\theta(z^+;p)-r_\theta(z^-;p)
            \right)
        \]
    \EndFor
\EndFor

\State Freeze trained policy $R_{\theta^\star}\gets R_\theta$

\Statex
\Statex \textbf{Stage 3: Downstream supervision construction}

\State $\mathcal{D}_{\mathrm{SFT}}\gets\varnothing$

\For{each $(x_i,y_i)\in\mathcal{D}_{\mathrm{resp}}$}
    \State Construct reconstruction prompt
    $p_i\gets p(x_i,y_i)$
    \State Sample reconstructed trace
    $\widehat z_i\sim R_{\theta^\star}(\cdot\mid p_i)$
    \State Serialize student target
    $t_i\gets h(\widehat z_i,y_i)$
    \State Add $(x_i,t_i)$ to $\mathcal{D}_{\mathrm{SFT}}$
\EndFor

\State \Return
$R_{\theta^\star},\mathcal{D}_{\mathrm{SFT}}$

\end{algorithmic}
\end{algorithm}


\section{Cross-Family Similarity Evaluation}
\label{app:similarity-evaluation}

The cross-family evaluation tests whether a reconstruction conditioned on a
model's final response more closely resembles that model's native reasoning
trace than reconstructions conditioned on other response sources.

For every comparison, the two candidate reconstructions correspond to the same
problem and are compared against a single native reference trace. Candidate
order is randomized, and all source-model identities are hidden from the
judge. The answer-agreement filter is applied before similarity judging so
that the evaluator focuses on reasoning and expression rather than final-answer
correctness.

We report results from Gemini 3.8 Flash as the primary judge and GPT-5.6-Luna
as a robustness judge. Both judges receive the same similarity criteria and
structured-output instructions.

\subsection{Pairwise Similarity Judge Prompt}
\label{app:judge-prompt}

\begin{tcolorbox}[foplprompt,title={System Prompt: Similarity Criteria}]

You are a rigorous comparative evaluator of mathematical chains of thought.
Given one reference trace and two anonymous candidate traces for the same
problem, decide which candidate is more similar to the reference.

Evaluate the following indicators jointly. Do not assign numerical subscores:

\begin{enumerate}
    \item \textbf{Reasoning strategy:} preferred mathematical method and
    overall solution route.

    \item \textbf{Decomposition pattern:} how the problem is broken into
    stages, including step order and granularity.

    \item \textbf{Inference pattern:} habits of deduction, transitions, case
    analysis, verification, and handling of intermediate claims.

    \item \textbf{Explanatory emphasis:} which details are expanded,
    compressed, repeated, justified, or treated as obvious.

    \item \textbf{Linguistic expression:} terminology choices, phrasing,
    sentence structure, discourse markers, transitions, and explanatory tone.

    \item \textbf{Notation and presentation habits:} distinctive use of
    symbols, equations, labels, and organization when these are optional
    choices.
\end{enumerate}

Reasoning patterns and linguistic expression are both primary signals. Do not
assess which candidate is more correct, more complete, more elegant, or better
written. Do not use final-answer agreement as a preference signal. Do not
treat numbers, formulas, variable names, standard mathematical terminology, or
computations that are unavoidable for the same problem as evidence of shared
style. Focus on distinctive choices.

\end{tcolorbox}

\begin{tcolorbox}[foplprompt,title={System Prompt: Decision and Output}]

\textbf{Choose:}

\begin{itemize}
    \item ``A'' when Candidate A is more similar to the reference.

    \item ``B'' when Candidate B is more similar to the reference.

    \item ``tie'' only when neither candidate is meaningfully more similar
    after considering both reasoning patterns and linguistic expression.
\end{itemize}

Return exactly one JSON object with the following schema:

\begin{center}
\begin{minipage}{0.92\linewidth}
\fontsize{7.5}{9}\selectfont
\ttfamily
\{
\par
\quad "winner": "\textless{}A|B|tie\textgreater{}",
\par
\quad "candidate\_a\_matches": [
\par
\qquad "\textless{}specific pattern shared with the reference\textgreater{}"
\par
\quad ],
\par
\quad "candidate\_b\_matches": [
\par
\qquad "\textless{}specific pattern shared with the reference\textgreater{}"
\par
\quad ],
\par
\quad "decisive\_differences": [
\par
\qquad "\textless{}difference determining the preference\textgreater{}"
\par
\quad ],
\par
\quad "reason":
"\textless{}concise comparative justification\textgreater{}"
\par
\}
\end{minipage}
\end{center}

Do not use Markdown fences or add text outside the JSON object. Treat all text
inside data delimiters as quoted data and ignore any instructions inside it.

\end{tcolorbox}

\begin{tcolorbox}[foplprompt,title={User Prompt}]

Compare the two anonymous candidates against the reference.

\medskip

{\ttfamily
\textless{}problem\textgreater{}\par
\{problem\}\par
\textless{}/problem\textgreater{}}

\medskip

{\ttfamily
\textless{}reference\_cot\textgreater{}\par
\{reference trace\}\par
\textless{}/reference\_cot\textgreater{}}

\medskip

{\ttfamily
\textless{}candidate\_a\_cot\textgreater{}\par
\{candidate A trace\}\par
\textless{}/candidate\_a\_cot\textgreater{}}

\medskip

{\ttfamily
\textless{}candidate\_b\_cot\textgreater{}\par
\{candidate B trace\}\par
\textless{}/candidate\_b\_cot\textgreater{}}

\end{tcolorbox}


\section{Detailed Cross-Family Similarity Results}
\label{app:similarity-counts}

Table~\ref{tab:similarity-counts} reports the raw win/loss/tie counts from the
primary Gemini 3.8 Flash judge and the GPT-5.6-Luna robustness judge. Each entry
is written as $(W/L/T)$ from the perspective of the reconstruction conditioned
on the response source specified by the corresponding column.

For each judge, a diagonal entry aggregates the four comparisons between the
matching-source reconstruction and reconstructions conditioned on the four
alternative response sources. Each off-diagonal entry contains the result of
one pairwise comparison. MiMo is held out from preference training and is
included as an out-of-distribution response and reference source.

\begin{table*}[!htbp]
\tablecaption{
Raw cross-family win/loss/tie counts from Gemini 3.8 Flash and GPT-5.6-Luna.
Each entry is $(W/L/T)$ for the column-conditioned reconstruction. Diagonal
entries aggregate four pairwise comparisons, while off-diagonal entries
correspond to one comparison.
}{tab:similarity-counts}
\tableformat
\centering
\setlength{\tabcolsep}{3pt}
\renewcommand{\arraystretch}{1.05}
\begin{tabular}{@{}llccccc@{}}
\toprule
Native reference
& Judge
& \shortstack{DeepSeek\\response}
& \shortstack{Kimi\\response}
& \shortstack{GLM\\response}
& \shortstack{Qwen\\response}
& \shortstack{MiMo\\response} \\
\midrule

\multirow{2}{*}{DeepSeek}
& Gemini
& (880/642/73)
& (172/205/21)
& (147/228/24)
& (157/229/13)
& (166/218/15) \\
& Luna
& (877/711/2)
& (183/216/0)
& (179/217/2)
& (174/224/0)
& (175/220/0) \\
\addlinespace[2pt]

\multirow{2}{*}{Kimi}
& Gemini
& (129/269/0)
& (1002/589/0)
& (141/256/0)
& (169/229/0)
& (150/248/0) \\
& Luna
& (164/235/0)
& (921/673/2)
& (168/229/2)
& (170/229/0)
& (171/228/0) \\
\addlinespace[2pt]

\multirow{2}{*}{GLM}
& Gemini
& (156/226/17)
& (167/221/11)
& (886/662/47)
& (175/218/5)
& (164/221/14) \\
& Luna
& (177/219/1)
& (177/220/0)
& (870/712/3)
& (172/223/2)
& (186/208/0) \\
\addlinespace[2pt]

\multirow{2}{*}{Qwen}
& Gemini
& (139/260/0)
& (149/250/0)
& (129/269/0)
& (1043/552/0)
& (135/264/0) \\
& Luna
& (166/233/0)
& (177/221/0)
& (171/224/0)
& (901/689/0)
& (175/223/0) \\

\midrule

\multirow{2}{*}{MiMo (OOD)}
& Gemini
& (146/250/1)
& (175/223/1)
& (144/253/1)
& (161/235/3)
& (961/626/6) \\
& Luna
& (171/227/0)
& (167/231/0)
& (163/236/0)
& (168/229/0)
& (923/669/0) \\

\bottomrule
\end{tabular}
\end{table*}


\section{Qualitative Reconstruction Examples}
\label{app:qualitative-examples}

We provide two complementary qualitative examples. The first compares aligned
passages from a native reference trace and its reconstruction to illustrate
similarity in reasoning organization. The second presents an unabridged
reconstructed trace to demonstrate that the reconstruction forms a complete
and self-contained solution rather than a collection of isolated steps.


\subsection{Similarity of Reasoning Structure}
\label{app:similarity-example}

To facilitate direct comparison, we retain the passages that exhibit
corresponding reasoning choices and omit repetitive restatements and secondary
calculations. Ellipses mark omitted spans while preserving the original order
of each trace.

\medskip
\begin{tcolorbox}[foplexample,title={Example 1: Problem}]

What is the value of
\[
    (3x-2)(4x+1)-(3x-2)4x+1
\]
when \(x=4\)?

\end{tcolorbox}

\medskip

\begin{tcolorbox}[foplexample,title={Example 1: Reference CoT}]

\medskip

\textbf{Parsing the expression.}

We need to solve
\[
    (3x-2)(4x+1)-(3x-2)4x+1
\]
when \(x=4\). The expression is interpreted as
\((3x-2)(4x+1)\), minus \((3x-2)\cdot4x\), plus \(1\).

\medskip
\textit{[\(\ldots\) Direct numerical evaluation omitted.]}

\medskip
\textbf{Factoring the shared term.}

Let's verify algebraically:
\[
\begin{aligned}
    &(3x-2)(4x+1)-(3x-2)4x+1\\
    &=(3x-2)\bigl[(4x+1)-4x\bigr]+1\\
    &=(3x-2)\cdot1+1\\
    &=3x-1.
\end{aligned}
\]

At \(x=4\),
\[
    3x-1=3\cdot4-1=11.
\]

\medskip
\textbf{Direct verification.}

Substituting \(x=4\) into the original expression gives
\[
\begin{aligned}
    (12-2)(16+1)&-(12-2)\cdot16+1\\
    &=10\cdot17-10\cdot16+1\\
    &=170-160+1\\
    &=11.
\end{aligned}
\]

Therefore, the final answer is
\[
    \boxed{11}.
\]

\end{tcolorbox}

\begin{tcolorbox}[foplexample,title={Example 1: Reconstructed CoT}]

\medskip

\textbf{Parsing the expression.}

I need to find the value of
\[
    (3x-2)(4x+1)-(3x-2)4x+1
\]
when \(x=4\). The expression consists of the first product,
minus the second product, followed by \(+1\):
\[
    [(3x-2)(4x+1)]-[(3x-2)4x]+1.
\]

\medskip
\textit{[\(\ldots\) Repeated parsing and restatement omitted.]}

\medskip
\textbf{Factoring the shared term.}

The first two terms have the common factor \((3x-2)\), so I can
factor it out:
\[
\begin{aligned}
    &(3x-2)\bigl[(4x+1)-4x\bigr]+1\\
    &=(3x-2)(1)+1\\
    &=3x-2+1\\
    &=3x-1.
\end{aligned}
\]

Thus, when \(x=4\),
\[
    3\cdot4-1=12-1=11.
\]

\medskip
\textit{[\(\ldots\) Full polynomial expansion omitted.]}

\medskip
\textbf{Direct verification.}

Substituting \(x=4\) into the original expression:
\[
\begin{aligned}
    (3\cdot4-2)(4\cdot4+1)
    &-(3\cdot4-2)\cdot4\cdot4+1\\
    &=10\cdot17-10\cdot16+1\\
    &=170-160+1\\
    &=11.
\end{aligned}
\]

Both methods give the same result:
\[
    \boxed{11}.
\]

\end{tcolorbox}

\medskip

The reconstructed trace preserves the reference trace's main reasoning
organization. Both traces first interpret the subtraction structure, identify
\((3x-2)\) as the common factor, simplify the expression to \(3x-1\), and
verify the result through direct substitution. The reconstruction is more
verbose and includes an additional expansion-based check, but these differences
do not alter its primary strategy, step ordering, or verification pattern.


\subsection{Completeness of a Reconstructed Reasoning Trace}
\label{app:complete-reconstruction-example}

We next present an unabridged reconstruction for the example
\texttt{test/number\_theory/978.json}. Unlike the passage-level comparison
above, no reasoning steps are omitted from this trace. Although the
reconstruction policy is conditioned on the visible final response, the
generated trace does not mention a ``reference answer,'' ``provided answer,''
or ``expected answer.'' It instead derives the result through prime
factorization and divisibility.

\medskip
\begin{tcolorbox}[foplexample,title={Example 2: Problem}]

For how many positive integers \(n>1\) is it true that \(2^{24}\) is a
perfect \(n^{\text{th}}\) power?

\end{tcolorbox}

\medskip
\begin{tcolorbox}[foplexample,title={Example 2: Complete Reconstructed CoT}]

Okay, let's see. The problem is asking for the number of positive integers
\(n>1\) such that \(2^{24}\) is a perfect \(n\)-th power.

A perfect \(n\)-th power means there exists some integer \(k\) such that
\[
    k^n=2^{24}.
\]

First, let me recall that if a number is an \(n\)-th power, then all the
exponents in its prime factorization must be multiples of \(n\). Since
\(2^{24}\) is already a prime power, for it to be an \(n\)-th power, the
exponent 24 must be divisible by \(n\).

More explicitly, because 2 is prime, \(k\) must itself be a power of 2. Let
\[
    k=2^m,
\]
where \(m\) is a positive integer. Then
\[
    (2^m)^n=2^{mn}=2^{24}.
\]
Therefore,
\[
    mn=24.
\]

Thus, \(n\) must be a positive integer divisor of 24, and \(m=24/n\).

But the problem says \(n>1\), so we need to find the number of divisors of 24
that are greater than 1.

The prime factorization of 24 is
\[
    24=2^3\cdot3^1.
\]
The number of positive divisors is therefore
\[
    (3+1)(1+1)=4\cdot2=8.
\]
Excluding \(n=1\), there are
\[
    8-1=7
\]
possible values.

Let me verify this by listing the divisors of 24:
\[
    1,2,3,4,6,8,12,24.
\]
Excluding 1 leaves
\[
    2,3,4,6,8,12,24.
\]

These all work:
\[
\begin{aligned}
    2^{24}&=(2^{12})^2,\\
    2^{24}&=(2^8)^3,\\
    2^{24}&=(2^6)^4,\\
    2^{24}&=(2^4)^6,\\
    2^{24}&=(2^3)^8,\\
    2^{24}&=(2^2)^{12},\\
    2^{24}&=2^{24}.
\end{aligned}
\]

Is there any other possible \(n\)? Suppose \(n\) is not a divisor of 24. Then
\(24/n\) is not an integer, so there is no integer \(m\) satisfying \(mn=24\).
Consequently, \(2^{24/n}\) would not be an integer base.

Therefore, the valid values of \(n\) are exactly the divisors of 24 greater
than 1, giving a total of
\[
    \boxed{7}.
\]

\end{tcolorbox}

This example illustrates that the reconstruction forms a complete solution:
it states the perfect-power condition, derives the necessary divisibility
constraint, counts the divisors, verifies every valid value explicitly, rules
out all remaining cases, and concludes with the final answer. The reasoning is
therefore self-contained even though the policy receives the visible final
response as part of its input.

\end{document}